\documentclass[graybox]{svmult}

\usepackage{mathptmx}       
\usepackage{helvet}         
\usepackage{courier}        
\usepackage{type1cm}        
\usepackage{makeidx}         
\usepackage{graphicx}        
\usepackage{multicol}        
\usepackage[bottom]{footmisc}
\usepackage{color}

\usepackage{xcolor}
\usepackage{float}

\makeindex             

\usepackage[T1]{fontenc}

\usepackage{url}

\usepackage{graphicx}
\usepackage[cmex10]{amsmath}
\usepackage{amsfonts}

\begin{document}

\title*{Adaptive Down-Sampling and Dimension Reduction in Elastic Kernel Machines for Efficient Recognition of Isolated Gestures}
\titlerunning{Adaptive Down-Sampling and Dimension Reduction in Elastic Kernel Machines}

\author{Pierre-Francois Marteau and Sylvie Gibet and Cl\'{e}ment Reverdy}
\institute{Pierre-Francois Marteau \at IRISA (UMR 6074),  Universit\'{e} de Bretagne Sud Campus de Tohannic, 56000 Vannes, France,
\email{firstname.name AT univ-ubs DOT fr}
\and Sylvie Gibet \at IRISA (UMR 6074),  Universit\'{e} de Bretagne Sud Campus de Tohannic, 56000 Vannes, France,
\email{firstname.name AT univ-ubs DOT fr}
\and Cl\'{e}ment Reverdy at IRISA (UMR 6074),  Universit\'{e} de Bretagne Sud Campus de Tohannic, 56000 Vannes, France,
\email{firstname.name AT univ-ubs DOT fr}
}

%
%
\maketitle

\begin{abstract} \\
In the scope of gestural action recognition, the size of the feature vector representing movements is in general quite large especially when full body movements are considered. Furthermore, this feature vector evolves during the movement performance so that a complete movement is fully represented by a matrix $M$ of size $DxT$, whose element $M_{i,j}$ represents the value of feature $i$ at timestamps $j$. Many studies have addressed dimensionality reduction considering only the size of the feature vector lying in $\mathbb{R}^D$ to reduce both the variability of gestural sequences expressed in the reduced space, and the computational complexity of their processing. In return, very few of these methods have explicitly addressed the dimensionality reduction along the time axis. Yet this is a major issue when considering the use of elastic distances which are characterized by a quadratic complexity along the time axis. We present in this paper an evaluation of straightforward approaches aiming at reducing the dimensionality of the matrix $M$ for each movement, leading to consider both the dimensionality reduction of the feature vector as well as its reduction along the time axis.
The dimensionality reduction of the feature vector is achieved by selecting remarkable joints in the skeleton performing the movement, basically the extremities of the articulatory chains composing the skeleton. The temporal dimensionality reduction is achieved using either a regular or adaptive down-sampling that seeks to minimize the reconstruction error of the movements.  Elastic and Euclidean kernels are then compared through support vector machine learning.
Two data sets that are widely referenced in the domain of human gesture recognition, and quite distinctive in terms of quality of motion capture, are used for the experimental assessment of the proposed approaches. On these data sets we experimentally show that it is feasible, and possibly desirable, to significantly reduce simultaneously the size of the feature vector and the number of skeleton frames to represent body movements while maintaining a very good recognition rate. The method proves to give satisfactory results at a level currently reached by state-of-the-art methods on these data sets. We experimentally show that the computational complexity reduction that is obtained makes this approach eligible for \textit{real-time} applications.

\end{abstract}

\section{Introduction}

Gesture recognition is a challenging task in the computer vision community with numerous applications using motion data such as interactive entertainment, human-machine interaction, automotive, or digital home. Recently, there is an increasing availability of large and heterogeneous motion captured data characterized by a various range of qualities, depending on the type and quality of the motion sensors. We thus separate (i) databases of high resolution and quality built from expensive capturing devices and requiring a particular expertise, (ii) and low resolution and noisier databases produced with cheap sensors that do not require any specific expertise.
Such databases open new challenges for assessing the robustness and generalization capabilities of gesture recognition algorithms on diversified  motion data sets. Besides the quality of recognition, the complexity of the algorithms and their computational cost is indeed a major issue, especially in the context of real-time interaction.

We address in this paper the recognition of isolated gestures from motion captured data. As motion data are generally represented by high-dimensional time series, many approaches have been developed to reduce their dimension, so that the recognition process is more efficient while being still accurate. Among them, low-dimensional embeddings of motion data have been proposed that enable to characterize and parameterize action sequences. 
Some of them are based on statistical descriptors \cite{Hussein2013}, rely on relevant meaningful trajectories \cite{OfliF2013}, or characterize the style\cite{Hussein2013}. In this paper we focus on dimensionality reduction along two complementary axes: the spatial axis representing the configuration of the skeleton at each frame, and the temporal axis representing the evolution over time of the skeletal joints trajectories. With such an approach, two main challenges are combined simultaneously:
\begin{itemize}
\item We use relevant trajectories (end-extremities) whose content may characterize complex actions;
\item Considering that the temporal variability is of primary importance when recognizing skilled actions and expressive motion, we apply an adaptive temporal down-sampling to reduce the complexity of elastic matching.
\end{itemize} 
These low-dimensional dual-based representations will be coupled with appropriate recognition algorithms that we expect to be more tractable and efficient.


Our recognition principle is based on a recent method that improves the performance of classical support vector machines when used with regularized elastic kernels dedicated to time series matching. Our objective is to show how the spatial and temporal dimensionality reductions, associated with such regularized elastic kernels significantly improve the efficiency, in terms of response time, of the algorithm while preserving the recognition rates.

The second section briefly presents the related works and state-of-the-art for isolated gesture recognition. In the third section we describe the nature of the motion data as well as the main pre-processing of the data. The fourth section gives the major keypoints of the method, positioning it in the context of multivariate sequential data classification. We present in the fifth part the evaluation of our algorithm carried out on two data sets with very distinct qualities and compare its performance with those obtained by some of the state-of-the-art methods. A final discussion is provided as well as some perspectives.

\section{Related work}
In human gesture recognition, there are two main challenges to be addressed: dimension reduction closely linked to feature descriptors, and recognition models which cover different aspects of dynamic modeling, statistical and machine learning approaches. In this section, we give a brief and non-exhaustive overview of the literature associated with each challenge.\\

{\bf Dimension reduction}

The problem of dimensionality reduction (also called manifold learning) can be addressed with the objective to find a low-dimensional space that best represents the variance of the data without loosing too much information.
Action descriptors have thus been defined for characterizing whole motion sequences, or punctual frames that need additional step of temporal modeling to achieve the recognition goal.

Numerous method are available to carry out such dimensionality reduction, the most popular being linear approaches such as  Principal Component Analysis (PCA, \cite{jolliffe1986}, ~\cite{Masoud2003}), Linear Discriminant Analysis (LDA, \cite{mclachlan2004}), or linear projections preserving locally neighborhoods (Locality Preserving Projection) \cite{He2003LPP}. Among non-linear approaches, Locally Linear Embeddings (LLE, \cite{Roweis2000}), Metric Multidimensional Scaling (MDS, \cite{KruskalWish1978}) and variants like Laplacian Eigenmap ~\cite{Belkin02}, Isomap~\cite{Tenenbaum2000} have been implemented to embed postures in low dimensional spaces in which a more efficient time warp (DTW, see section \ref{sec:DTW}) algorithm can be used to classify movements. An extension of this method, called ST-Isomap, considers temporal relationships in local neighborhoods that can be propagated globally via a shortest-path mechanism~\cite{Jenkins:ICML04}. Models based on Gaussian processes with latent variables are also largely used, for instance a hierarchical version has been recently exploited for gesture recognition \cite{Han:2010}.


Other methods define discriminative features that best classify motion classes. This is the case in the work of \cite{YuAggarwal2009} that  reduces the motion data to only five end-extremities of the skeleton (two feet, two hands and the head), thus giving some meaningful insight of the motion related to the action task. \cite{Fothergill:2012} and \cite{Zhao2012} have in particular applied random forests to recognize actions, using a Kinect sensor, while \cite{OfliF2013} recently proposed to automatically select the most informative skeletal joints to explain the current action.
In the same line, \cite{Hussein2013} use the covariance matrix for skeleton joint locations over time as a discriminative descriptor to characterize a movement sequence. Multiple covariance matrices are deployed over sub-sequences in a hierarchical fashion in order to encode the relationship between joint movement and time.
In \cite{Li2010} a simple bag of 3D points is used to represent and recognize gestural action. Similarly, in \cite{Wang2012}, \textit{actionlets} are defined from Fourrier coefficients to characterize the most discriminative joints. 
Finally, it can be mentioned, among many existing applications that address the use of elastic distances into a recognition process, the recent work described in \cite{Sempena2011},  as well as the hardware acceleration proposed in \cite{Hussain2012}. However, to our knowledge, no work exploiting this type of distance has directly studied the question of data reduction along the time axis.

{\bf Gesture recognition}
Recognition methods essentially aim at modeling the dynamics of gestures. 
Some approaches, based on linear dynamic models  \cite{Veeraraghavan2004}, have used autoregressive (AR) and autoregressive moving-average (ARMA) models to characterize the kinematics of movements, while other approaches, based on nonlinear dynamic models \cite{Bissacco2007}, have developed movement analysis and recognition scheme based on dynamical models controlled by Gaussian processes. \cite{Mitra:2007} propose a synthesis of the major gesture recognition approaches relying on Hidden Markov Models (HMM). Histograms of oriented 4D normals have also been proposed in \cite{Oreifej2013} for the recognition of gestural actions from sequences of depth images. 
\cite{Wang2006} have exploited conditional random fields to model joint dependencies and thus increase the discrimination of HMM-like models. Recurrent neural network models have also been used \cite{Martens2011}; among them, conditional restricted Boltzman's machines \cite{Larochelle2012} have been studied recently in the context of motion captured data modeling and classification.

In this paper, we propose a new representation of human actions that results from a dual-based reduction method that occurs both spatially and temporally. We couple this representation to a SVM classification method associated with regularized elastic kernels. 


\section{Motion representation}
We are working on isolated human motions acquired through motion captured databases. In recent years, there is an increasing availability of these databases, some of them being captured by high resolution devices (infrared marker-tracking system), such as those provided by CMU \cite{cmu2003}, HDM05 \cite{HDM05}, and other ones captured by low-cost devices, such as MSRAction3D captured with the Microsoft Kinect system. With the first type of sensors, the acquisition process is expensive, as it necessitates for capturing skilled motion with a good accuracy several cameras with many markers located on an actor, and a post-processing pipeline which is costly in time and expertise. Instead, with the second type of sensors, the data acquisition is affordable and necessitates less time and expertise, but with a loss of accuracy which remains acceptable for some tasks.

After the recording, the captured data is filtered and reconstructed so that to eliminate most of the noise, data loss and labelling errors (markers inversion), and the output of this acquisition pipeline is generally 
a set of 3D-trajectories of the skeleton joints determined from the positions of markers. 
This kind of data is inherently noisy, mainly due to the quality of the sensors and the acquisition process, but also to approximations made during the reconstruction process. The modeling of the skeleton is indeed subject to some variation for different reasons: in particular the markers being positioned on cloths or on the skin of the actor's body, they can move during the capturing process, also the determination of the segment lengths, of the joints' centers and their associated rotation axis are not trivial and lead to modeling errors. To overcome these difficulties, the skeleton model is obtained through an optimization process such as the ones described in \cite{AguiarTS06}, \cite{Obrien:2000}, or \cite{Shotton:2011}. The techniques based on a skeleton model hence convert 3D sensor data into Cartesian or angular coordinates that define the state of the joints over time  with various accuracies.


\begin{figure*}[ht!]
  \centering
  \begin{tabular}{cc}
	\includegraphics[scale=.25]{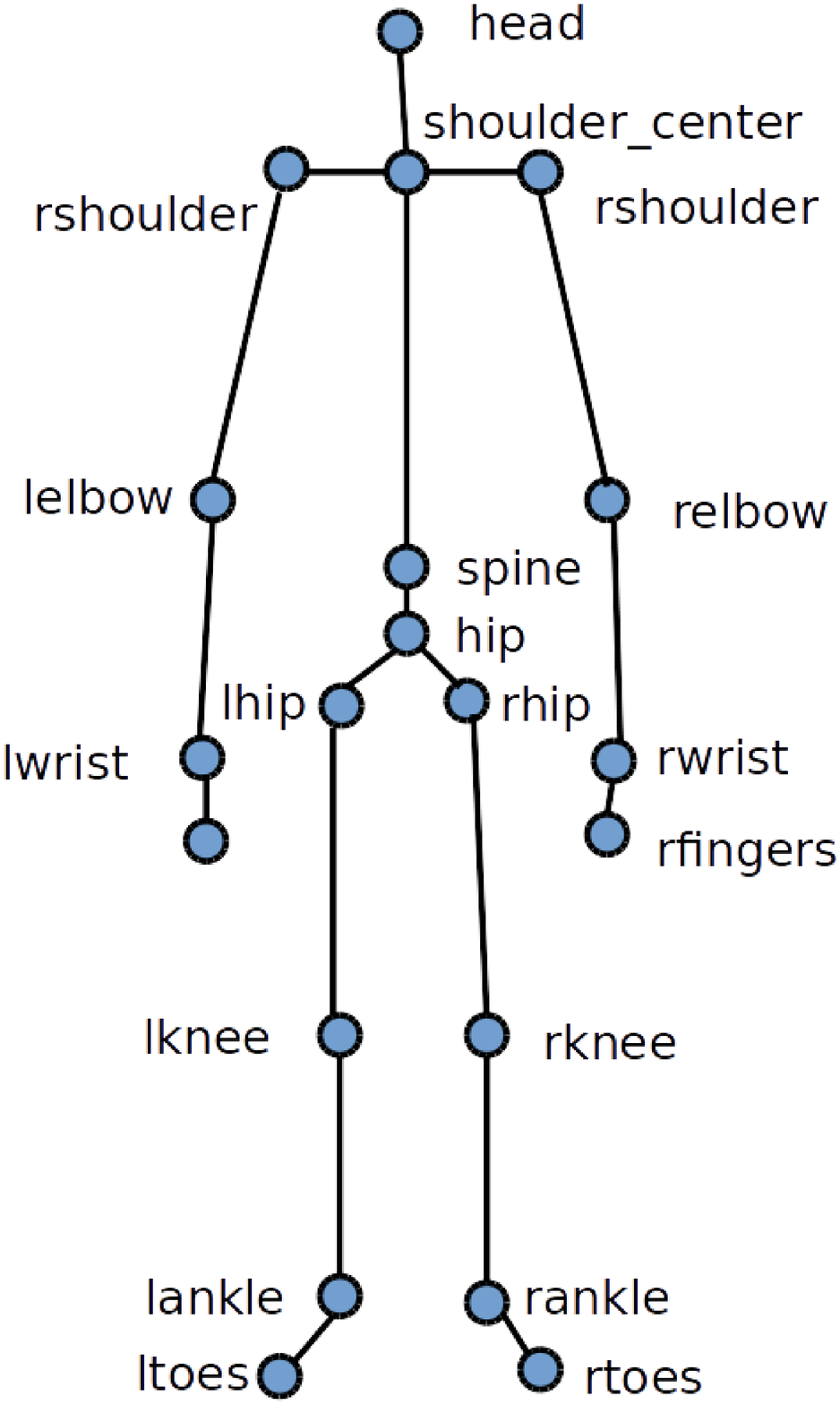}&
    \includegraphics[height=70mm]{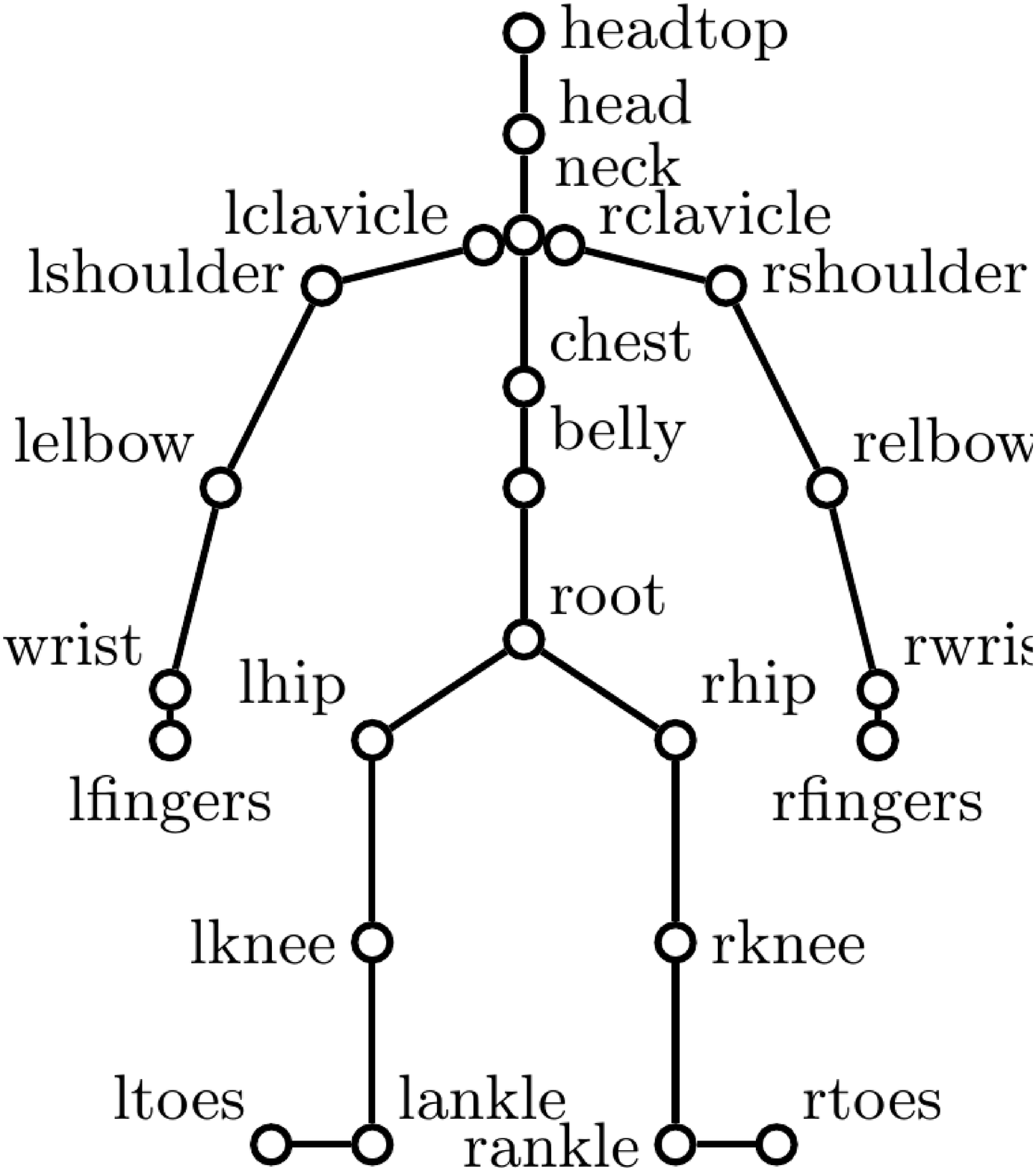}
  \end{tabular}
  \caption{Examples of skeletons reconstructed from motion data captured from the Kinect (left) and from the Vicon-MX device used by the Max Planck Institute (right).}
\label{fig:skels}
\end{figure*}

Figure \ref{fig:skels} presents two skeletons reconstructed from two very distinct capture systems. On the right, the skeleton is reconstructed from data acquired via the Microsoft Kinect, on the left from the Vicon-MX device used by the Max Planck Institute to produce the  HDM05 datasets.

Thus, any skeletal-based model can be represented by a hierarchical tree-like structure composed of rigid segments (bones) connected by joints. A motion can be defined by a sequence of postures over time, formalized as a multivariate state vector describing a trajectory over time, i.e. a time series:
$\{Y_t \in\mathbf{R}^{k}\}^{T}_{1} = [Y_1 , . . . , Y_T]$, where the $k$ spatial dimension ($k=3\cdot N$, with N the number of joints) typically varies between $20$ and $100$ according to the capture devices and the considered task. 
As this state vector is obviously not composed of independent scalar dimensions, the spatio-temporal encoded redundancies open solutions for dimension reduction as well as noise reduction approaches. This is particularly relevant for motion recognition, as the objective is to aim at improving computation time and error rate.

\section{Dimension reduction of motion capture data}
Using elastic distances or kernels for recognition problems has proved to be very accurate and efficient. However, one of the main difficulty of such methods with time series data is to deal with their computational cost, in general quadratic with the length of the time series and linear with the \textit{spatial} dimension characterized by the number of degrees-of-freedom. This high computational complexity is potentially limiting their use, especially when large amounts of data have to be processed, or when real-time constraint is required. We therefore expect that a dual dimensionality reduction, both on the time and spatial axis is particularly relevant, especially when associated to these techniques. Hence, for motion recognition using elastic distances, we propose to show that there exists a spatio-temporal redundancy in motion data that can be exploited.

Considering the quite rich literature on motion recognition, it appears that while some studies have shown success with dimensionality reduction on the spatial axis, very few have directly addressed a dimensionality reduction along the time axis 
, and much less work by combining the two approaches. 
\cite{Keogh:2000} has however proposed a temporal sub-sampling associated with dynamic time warping in the context of time series mining, followed later by \cite{MarteauTWED09}. We address herein after these two lines of research.

\subsection{Dimension reduction along the spatial axis}
Dimension reduction (or manifold learning), is the process consisting of mapping high dimensional data to representations lying in a lower-dimensional manifold. 
This can be interpreted as mapping the data into a space characterized by a smaller dimension, from which most of the variability of the data can be reproduced.

We consider in this paper a more direct approach based on the knowledge of the mechanism underlying 
the production of motion data 
and the way human beings perceive and discriminate body movements. 
We make the assumption that the perception of human motion is better achieved in the so-called \emph{task-space} represented by a selection of significant 3D joint trajectories. This hypothesis is supported by \cite{Giese2008} who show that visual perception of body motion closely reflects physical similarities between joint trajectories.  
This is also consistent with the motor theory of motion perception presented in \cite{Gibet2011}. 
Besides, we may reasonably accept that these joint trajectories embed sufficient discriminative information as inverse kinematics (widely used in computer animation and robotics) has shown to be very efficient and robust to reconstruct the whole skeleton movement from the knowledge  of the end effector trajectories (hands, feet, head), possibly with the additional knowledge of mid-articulated joints trajectories (such as elbows and knees) and constraints. Hence a straightforward approach consists in constructing a motion descriptor that discards all joints information but the 3D positions of the mid and end effectors extremities. We thus select the 3D positions for the two wrists, the two ankles (the fingers and toes markers are less reliable in general), the two elbows, the two knees and the head. This leads to a time-varying descriptor lying in a 18D space, while a full body descriptor is embedded in a 60D, space for the Kinect sensor, significantly more for vicon settings in general.
 
\subsection{Dimension reduction along the time axis}

The straightforward approach we are developing to explicitly reduce dimensionality along the time axis consists in sub-sampling the movement data such that each motion trajectories takes the form of a fixed-size sequence of $L$ skeletal postures. Then it becomes easy to perform a classification or recognition task by using elastic kernel machines on such fixed-size sequences. With such an approach, performance rates depend on 
the chosen degree of sub-sampling. 
This approach seems coarse since long sequences are characterized with the same number of skeletal poses than short sequences. For very short sequences, whose lengths are shorter than $L$, if any, we over-sample the sequence to meet the fixed-size requirement. But we consider this case as very marginal since we seek a sub-sampling rate much lower than the average sequence of movement length. 
In the following, we will experiment and compare uniform and adaptive down-sampling.

\subsubsection{Uniform down-sampling}
In order to explicitly reduce dimensionality along the time axis, our first straightforward approach here consists in down-sampling the motion data so that each motion trajectory takes the form of a fixed-size sequence of $L$ skeletal postures, evenly distributed along the time axis. We refer to this approach as uniform down-sampling (UDS).

\subsubsection{Adaptive down-sampling}
The second approach is the so-called adaptive down-sampling (ADS) approach. Similarly to UDS, each motion trajectory takes the form of a fixed-size sequence of $L$ skeletal postures, but these postures are not evenly distributed anymore along the time axis. They are selected such as to minimize a trajectory reconstruction criteria. Basically we follow the previous work by \cite{MarteauGibet2006}. A data modeling approach is used to handle the adaptive sampling of the $\{ Y_t\}$ multidimensional and discrete time series. More precisely, we are seeking an approximation $Y_{\theta^*}$ of $Y$ such as:
\begin{equation}
\theta^* = \underset{\theta}{ArgMin} ( E (\{Y_t\}, \{Y_\theta,t\} ))
\label{eq:ADS}
\end{equation}

\begin{figure}[h]
  \centering
  \begin{tabular}{cc}
  \includegraphics[scale=.25]{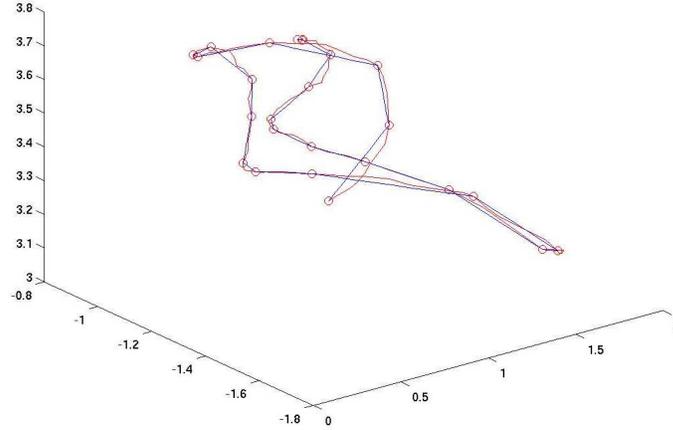} \\
  \end{tabular}
  \caption{Trajectory of the human wrist in the 3D Cartesian space adaptively dow-sampled with the localization of the 25 selected samples (red circles): motion capture  data (blue) and reconstructed data by linear interpolation (red).}
\label{fig:pla}
\end{figure}

where $E$ is the RMS error between $Y$ and $Y_\theta$ selected among the set of piecewise linear approximations defined from $Y$. Since the optimal solution of the optimization problem defined by Eq. \ref{eq:ADS} is $O(L.n^2)$, where $n$ is the length of the input time series, we adopt a near to optimal solution as developed in \cite{MarteauPAA09} whose complexity is $O(n)$. As an example, in figure \ref{fig:pla} the human wrist 3D trajectory is down-sampled using 25 samples positioned along the trajectory  by minimizing the piecewise linear approximation.

\section{Elastic kernels and their regularization}    
\label{sec:DTW}
\textbf{Dynamic Time Warping} (DTW), \cite{TWED:Velichko70}, \cite{TWED:Sakoe71}, by far the most used elastic measure, is defined as 

\begin{eqnarray}
\label{Eq.2}
 d_{dtw}(X_p,Y_q)&= &d_{E}^{2}(x(p),y(q))  \\
  &+&\text{Min} 
   \left\{
   \begin{array}{ll}
     d_{dtw}(X_{p-1},Y_{q}) & sup \nonumber\\ 
     d_{dtw}(X_{p-1},Y_{q-1}) & sub	\nonumber	 \\
     d_{dtw}(X_{p},Y_{q-1}) & ins \nonumber \\
   \end{array}
   \right.
\end{eqnarray}
where $d_{E}(x(p),y(q)$ is the Euclidean distance (possibly the square of the Euclidean distance) defined on $\mathbb{R}^k$ between the two postures in sequences $X$ and $Y$ taken at times $p$ and $q$ respectively. 

When performed by a support vector machine (SVM) model, the optimization problem inherent to this type of learning algorithm is no longer quadratic. 
Moreover, the convergence towards the \textit{optimorum} is no longer guaranteed, which, depending on the complexity of the task may be considered as detrimental.
 
Besides the fact that the DTW measure does not respect the triangle inequality, it is furthermore not possible to directly define a positive definite kernel from it. Hence, the optimization problem, inherent to the learning of a kernel machine, is no longer quadratic which could, at least on some tasks, be a source of limitation.\\

\textbf{Regularized DTW}: recent works  \cite{Cuturi07},  \cite{Marteau2014} allowed to propose new guidelines to regularize kernels constructed from elastic measures such as DTW. A simple instance of such regularized kernel, derived from \cite{Marteau2014} for time series of equal length, takes the following form, which relies on two recursive terms :

\begin{align}
\label{Eq.MEREDK}
\begin{array}{ll}
K_{rdtw}(X_{p},Y_{q})=K^{xy}_{rdtw}(X_{p}, Y_{q})+K^{xx}_{rdtw}(X_{p},Y_{q}) \\
\\
K^{xy}_{dtw}(X_{p},Y_{q}) = \frac{1}{3}e^{-\nu d_{E}^{2}(x(p),y(q))}  \\
   \sum \left\{
   \begin{array}{ll}
    h(p-1,q)K^{xy}_{rdtw}(X_{p-1},Y_{q}) \\
   h(p-1,q-1) K^{xy}_{rdtw}(X_{p-1},Y_{q-1})  \\
    h(p,q-1)K^{xy}_{rdtw}(X_{p},Y_{q-1}) \\
   \end{array}
   \right.\\
\\
   K^{xx}_{rdtw}(X_{p},Y_{q}) = \frac{1}{3} \\
   \sum \left\{
   \begin{array}{ll}
    h(p-1,q) K^{xx}_{rdtw}(X_{p-1},Y_{q})e^{-\nu d_{E}^{2}(x(p),y(p))}  \\
    \Delta_{p,q} h(p,q)K^{xx}_{rdtw}(X_{p-1},Y_{q-1})e^{-\nu d_{E}^{2}(x(p),y(q))}   \\
    h(p,q-1)K^{xx}_{rdtw}(X_{p},Y_{q-1})e^{-\nu d_{E}^{2}(x(q),y(q))} \\
   \end{array}
   \right.\\
  \end{array}
\end{align}
where $\Delta_{p,q}$ is the Kronecker's symbol, $\nu \in \mathbb{R}^{+}$ is a \textit{stiffness} parameter which weights the local contributions, i.e. the distances between locally aligned positions, and $d_E(.,.)$ is a distance defined on $\mathbb{R}^{k}$. 

The initialization is simply $K^{xy}_{rdtw}(X_{0},Y_{0}) = K^{xx}_{rdtw} (X_{0},Y_{0}) = 1$.\\

The main idea behind this line of regularization is to replace the operators $\min$ and $\max$ (which prevent the symmetrization of the kernel) by a summation operator ($\sum$). This leads to consider, not only the best possible alignment, but also all the best (or nearly the best) paths by summing up their overall cost. The parameter $\nu$ is used to control what we call nearly-the-best alignment, thus penalizing more or less alignments too far from the optimal ones. This parameter can be easily optimized through a cross-validation. \\

\subsection{Normalization}
As $K_{rdtw}$ evaluates the sum on all possible alignment paths of the products of local alignment costs  $e^{- d_{E}^{2}(x(p),y(p))/(2.\sigma^2)} \le 1$, its values can be very small depending on the size of the time series and the selected value for $\sigma$. Hence, $K_{DTW}$ values tend to $0$ when $\sigma$ tends towards $0$, except when  the two compared time series are identical (the corresponding Gram matrix suffers from a diagonal dominance problem). As proposed in \cite{Marteau2014}, a manner to avoid numerical troubles consists in using the following \textit{normalized} kernel:
\[
\tilde{K}_{rdtw}(.,.)=exp\left(\alpha\frac{log(K_{rdtw}(.,.))-log(min(K_{rdtw}))}{log(max(K_{rdtw})) - min(K_{rdtw}))} \right)
\]
where $max(K_{rdtw})$ and $min(K_{rdtw})$ respectively are the max and min values taken by the kernel on the learning data set and $\alpha > 0$ a positive constant ($\alpha=1$ by default). 
If we forget the proportionality constant, this leads to take the kernel $K_{rdtw}$ at a power $\tau=\alpha/(log(max(K_{rdtw})) -log(min(K_{rdtw})))$, which shows that the normalized kernel  $\tilde{K}_{rdtw} \propto K_{rdtw}^\tau$ is still positive definite (\cite{BergChristensenRessel84}, Proposition 2.7).\\

We consider in this paper the non definite exponential kernel (Gaussian or Radial Basis Function (RBF) types) $K_{dtw}= e^{- d_{dtw}(.,.)/(2.\sigma^2)}$ constructed directly from the  elastic measures $d_{dtw}$, the normalized regularized elastic kernel $K_{rdtw}^\tau$,  and the non-elastic kernel obtained from the Euclidean distance \footnote {The Euclidean distance is usable only because a fixed number of skeletal positions is considered to characterize each movement, and this, irrespectively of their initial length}, i.e., $K_{E}(.,.)= e^{- d_{E}^2(.,.)/\sigma}$. 

%
%

\section{Experimentation}


To estimate the robustness of the proposed approaches, we evaluate them on two motion capture databases of opposite quality, the first one, called  \emph{HDM05}, developed at the Max Planck Institute, the other one, called \emph{MSR-Action3D}, at Microsoft research laboratories. \\

\textbf{HDM05 data set} \cite{HDM05} consists of data captured at 120Hz by a Vicon MX system composed of a set of reflective optical markers followed by six high-definition cameras and configured to record data at 120hz. 
The movement sequences are segmented and transformed into sequences of skeletal poses consisting of N = $31$ joints, each associated to a 3D position $(x, y, z)$. In practice the position of the root of the skeleton (located near its center of mass) and its orientation serving as referential coordinates, only the relative positions of the remaining 30 joints are used, which leads to represent each position by a vector $Y_T \in \mathbb{ R}^{k}$ , with $k=90$. We consider two recognition/classification tasks:  HDM05-1 and HDM05-2 that are respectively those proposed in \cite{OfliF2012} (also exploited in the work of \cite{Hussein2013}) and \cite{OfliF2013}. For both tasks, three subjects are involved during learning and two separate subjects are involved during testing. For task HDM05-1, 11 gestural actions are processed: \textit{\{deposit floor, elbow to knee, grab high, hop both legs, jog, kick forward, lie down floor, rotate both arms backward, sneak, squat, and throw basketball\}}. This constitutes 249 motion sequences. For task HDM05-2 , the subjects are the same, but five additional gestural actions are considered in addition to the previous 11: \textit{\{jump, jumping jacks, throw, sit down, and stand up\}}. For this task, the data set includes 393 movement sequences in total.
For both tests, the lengths of the gestural sequences are between 56 and 901 postures (corresponding to a movement duration between 0.5-7.5 sec.) . \\
 
\textbf{MSR-Action3D data set}: This database \cite{Li2010} has recently been developed to provide a Kinect data \textit{benchmark}. It consists of 3D depth image sequences (\textit{depth map}) captured by the Microsoft Kinect sensor. It
contains 20 typical interaction gestures with a game console that are labeled as follows \textit{high arm wave, horizontal arm wave, hammer, hand catch, forward punch, high throw, draw x, draw tick, draw circle, hand clap, two hand wave, side-boxing, bend, forward kick, side kick, jogging, tennis swing, tennis serve, golf swing,  pickup \& throw} . Each action was carried out by 10 subjects facing the camera, 2 or 3 times. This data set includes 567 motion sequences whose lengths vary from 14 to 76 skeletal poses. The 3D images of size $640 \times 480$  were captured at a frequency of 15hz.  From each 3D image a skeletal posture has been extracted with $N=20$ joints, each one being characterized by three coordinates. As for the previous data set, we characterize postures relatively to the referential coordinates located at the root of the skeleton, which leads to represent each posture by a vector $Y_t \in \mathbb{R}^{k}$, with $k=3 \times 19 = 57$.
The task is to provide a cross-validation on the subjects, i.e. 5 subjects participating in learning and 5 subjects participating in testing, considering all possible configurations which represent 252 learning/testing pairs in total.\\

Hence, we perform three classification tasks: HDM05-1, HDM05-2 and MSRAction3D, with or without a spatial dimensionality reduction while  simultaneously considering a down-sampling on the time axis:
\begin{itemize}

\item The spatial dimensionality reduction is obtained by constructing a frame (skeletal pose) descriptor composed only with the end-effector trajectories in 3D (EED) comparatively to a full-body descriptor (FBD) that integrates all the joints trajectories that compose the skeleton. The FBD rests in a 90D space for HDM05 and in a 60D space for MSRAction3D.  The EED rests in a 24D space (3D positions for 2 elbows, 2 hands, two knees and two feet) for the three tasks, leading to a data compression of 73\% for HDM05 and 55\% for MSRAction3D.\\

\item The dimensionality reduction on the time axis is obtained through either a uniform down-sampling (UDS) or an adaptive down-sampling (ADS) based on a piecewise approximation of the FBD or EED trajectories.  The number of skeletal poses varies from 5 to 30 for each trajectories, leading to an average data compression of 97\% for HDM05 and 74\% for MSRAction3D. 
\end{itemize}

\begin{figure}[H]
  \centering
  \begin{tabular}{cc}
	\includegraphics[scale=.5]{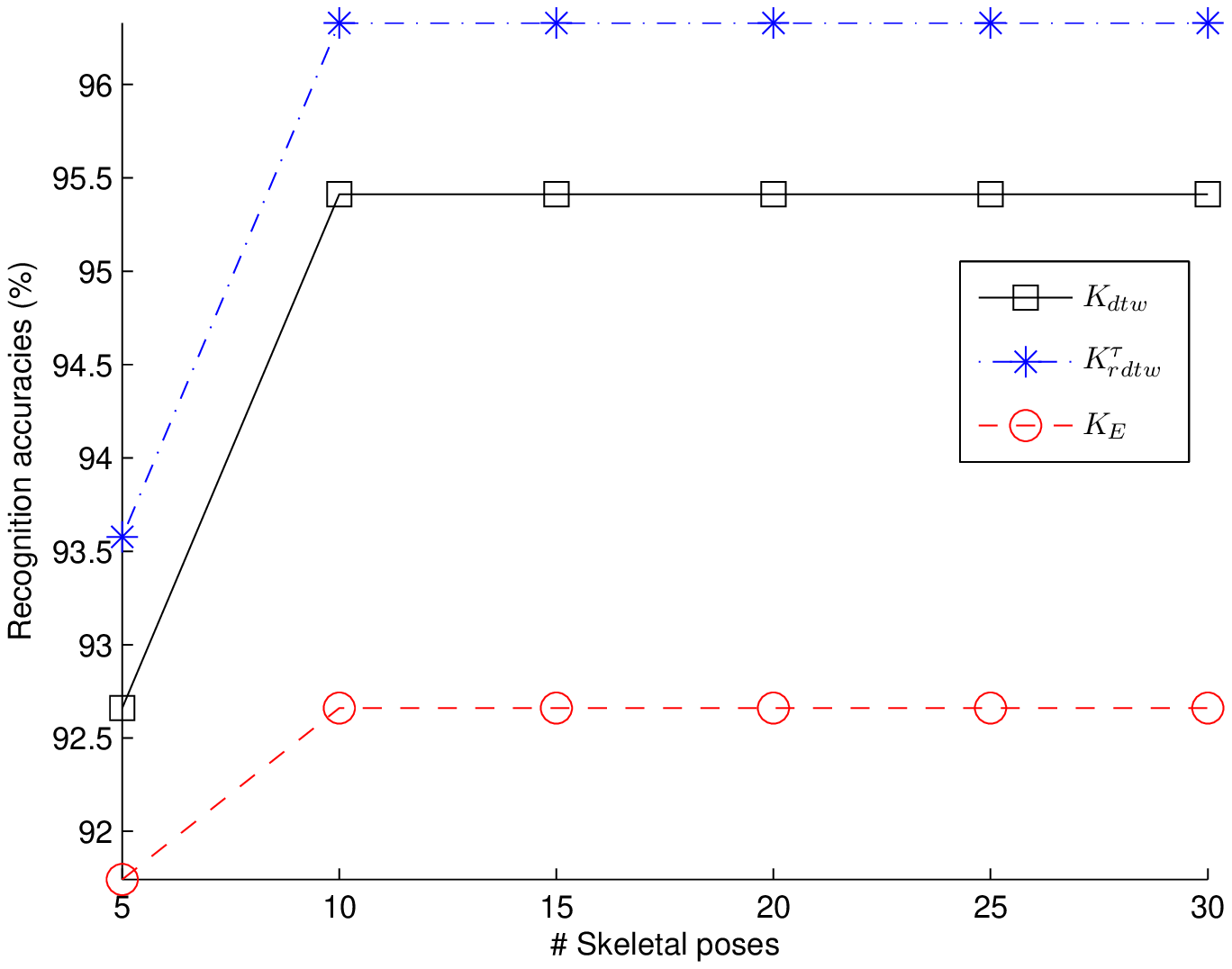} \\
    \includegraphics[scale=.5]{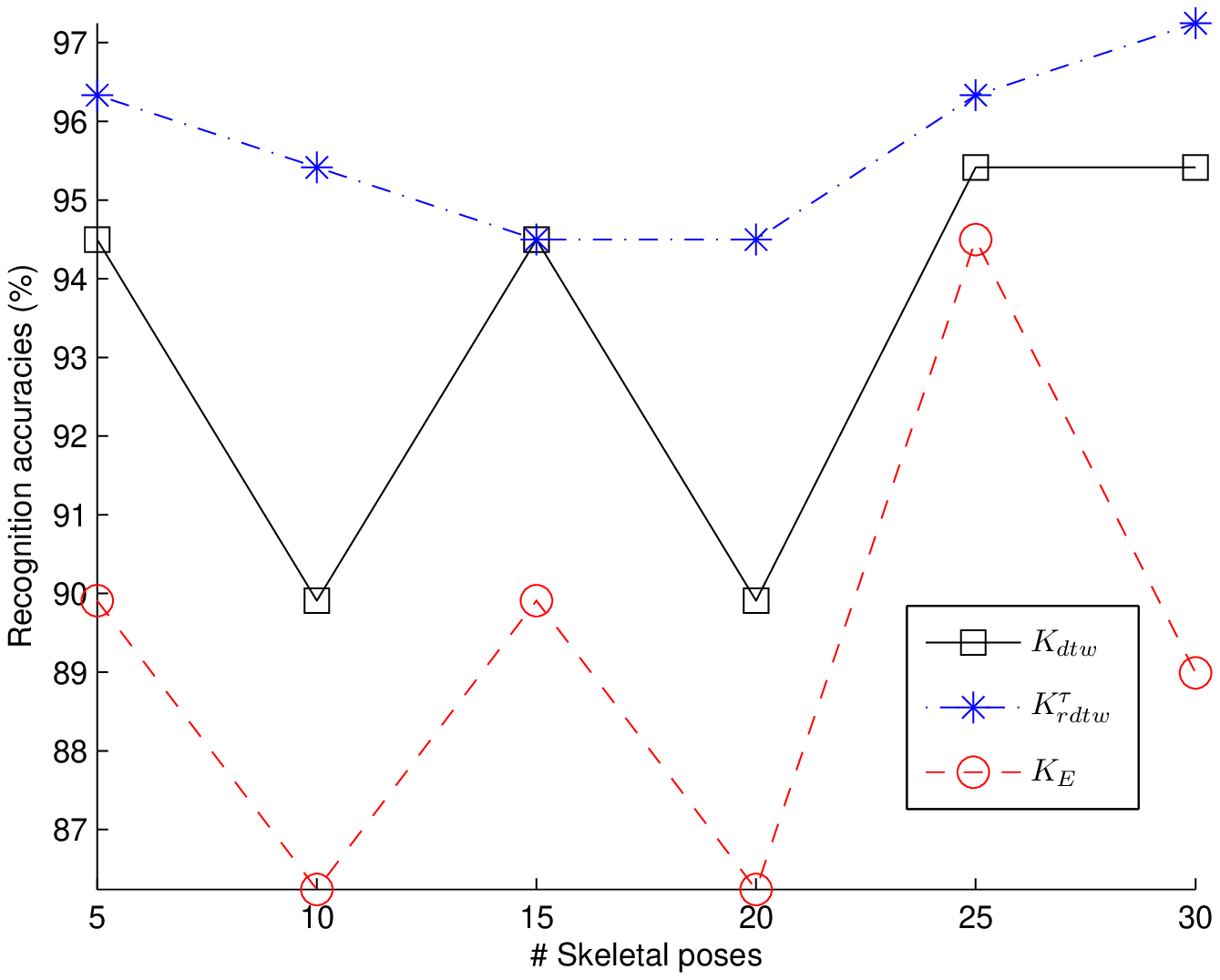} \\
    \includegraphics[scale=.5]{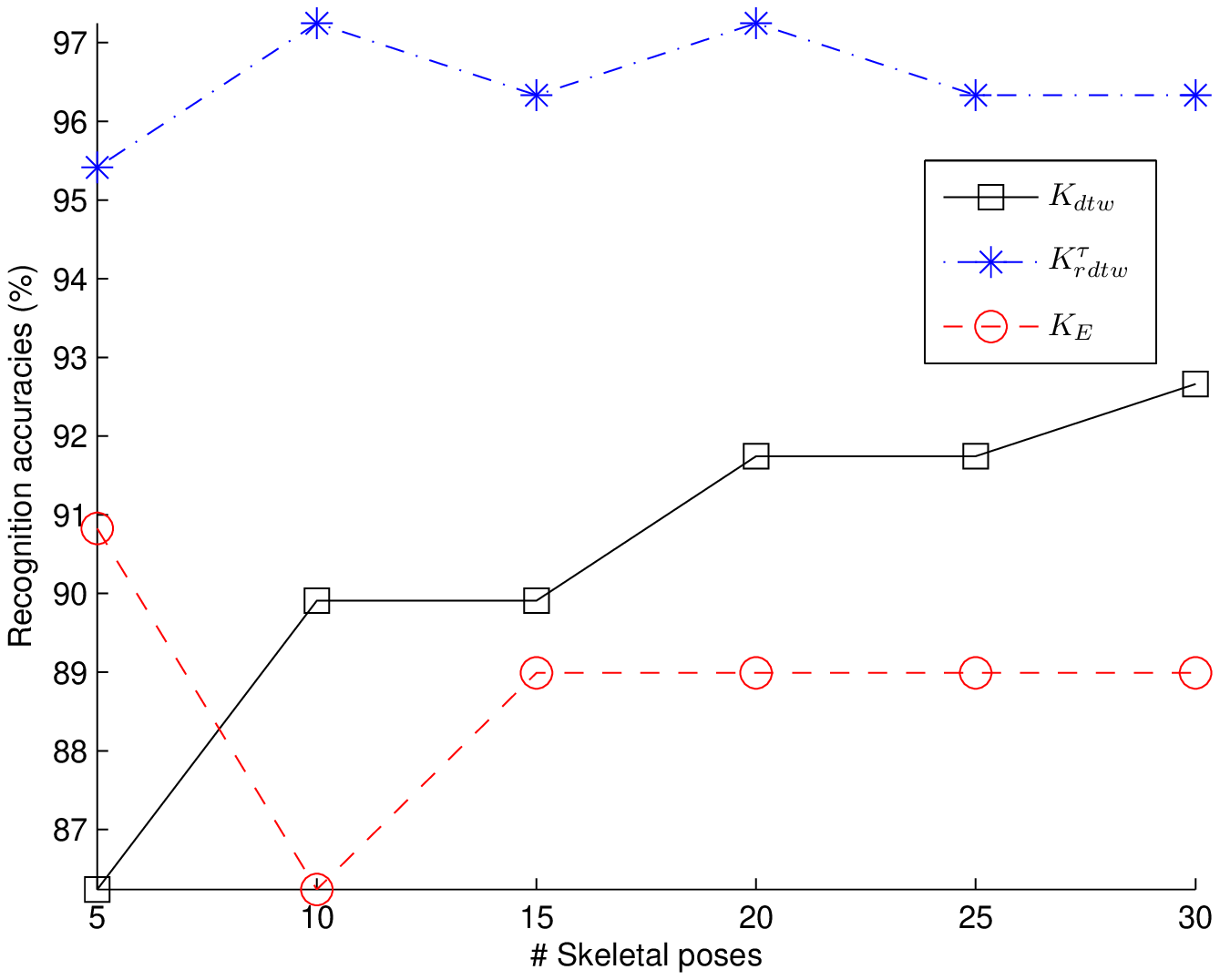}
  \end{tabular}
  \caption{Classification accuracies for HDM05-1 task 
  as defined in \cite{OfliF2012} 
  uniform down-sampling, full body (top), adaptive down-sampling, full body (middle), adaptive down-sampling, end effector extremities (bottom), when the number of skeletal poses varies: $K_{E}$ (red, circle, dash),  $K_{dtw}$ (black, square, plain), $K_{rdtw}^\tau$ (blue, star, dotted). }
\label{fig:HDM05Ofli2012}
\end{figure}

\begin{figure}[H]
  \centering
  \begin{tabular}{cc}
	\includegraphics[scale=.55]{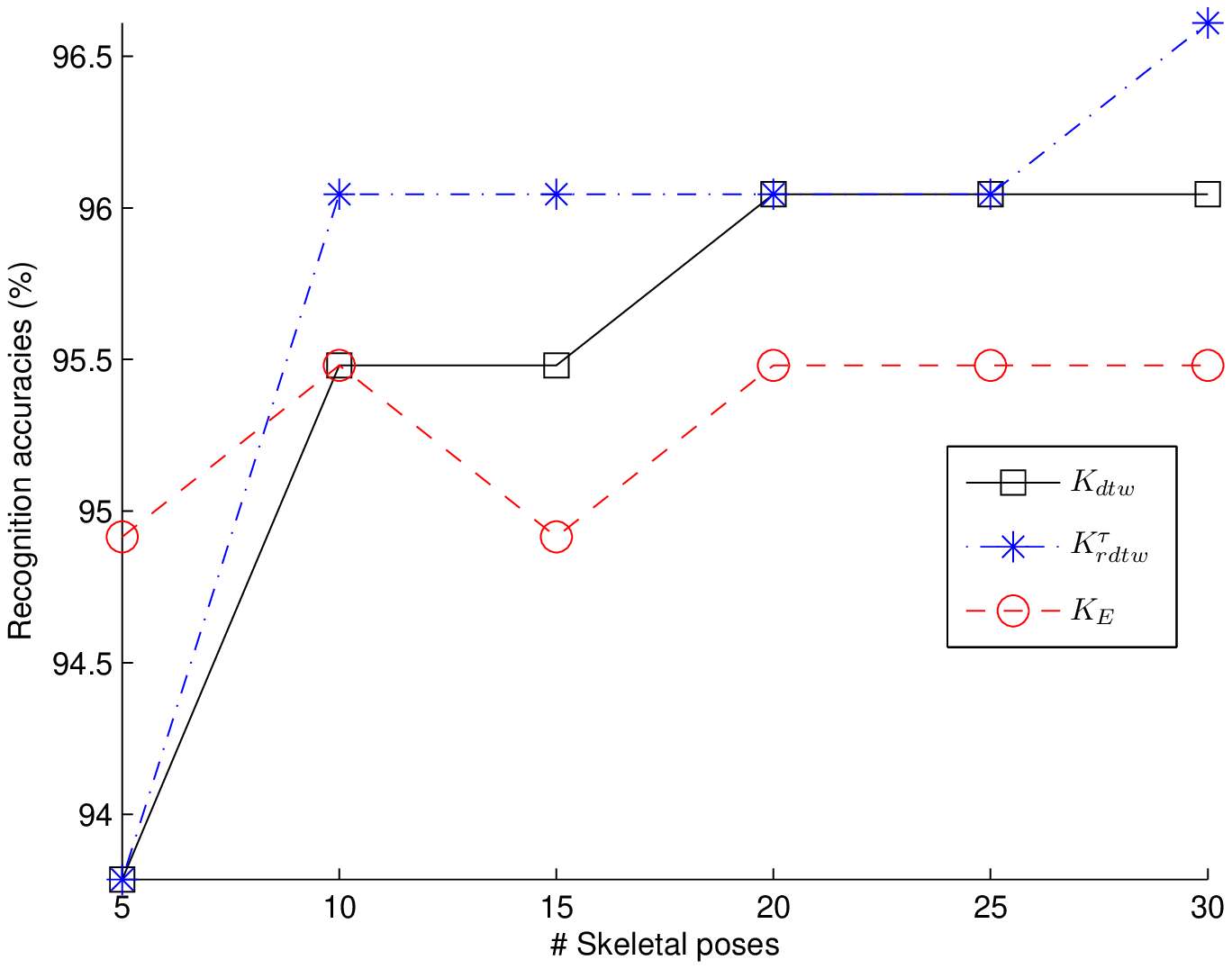} \\
    \includegraphics[scale=.55]{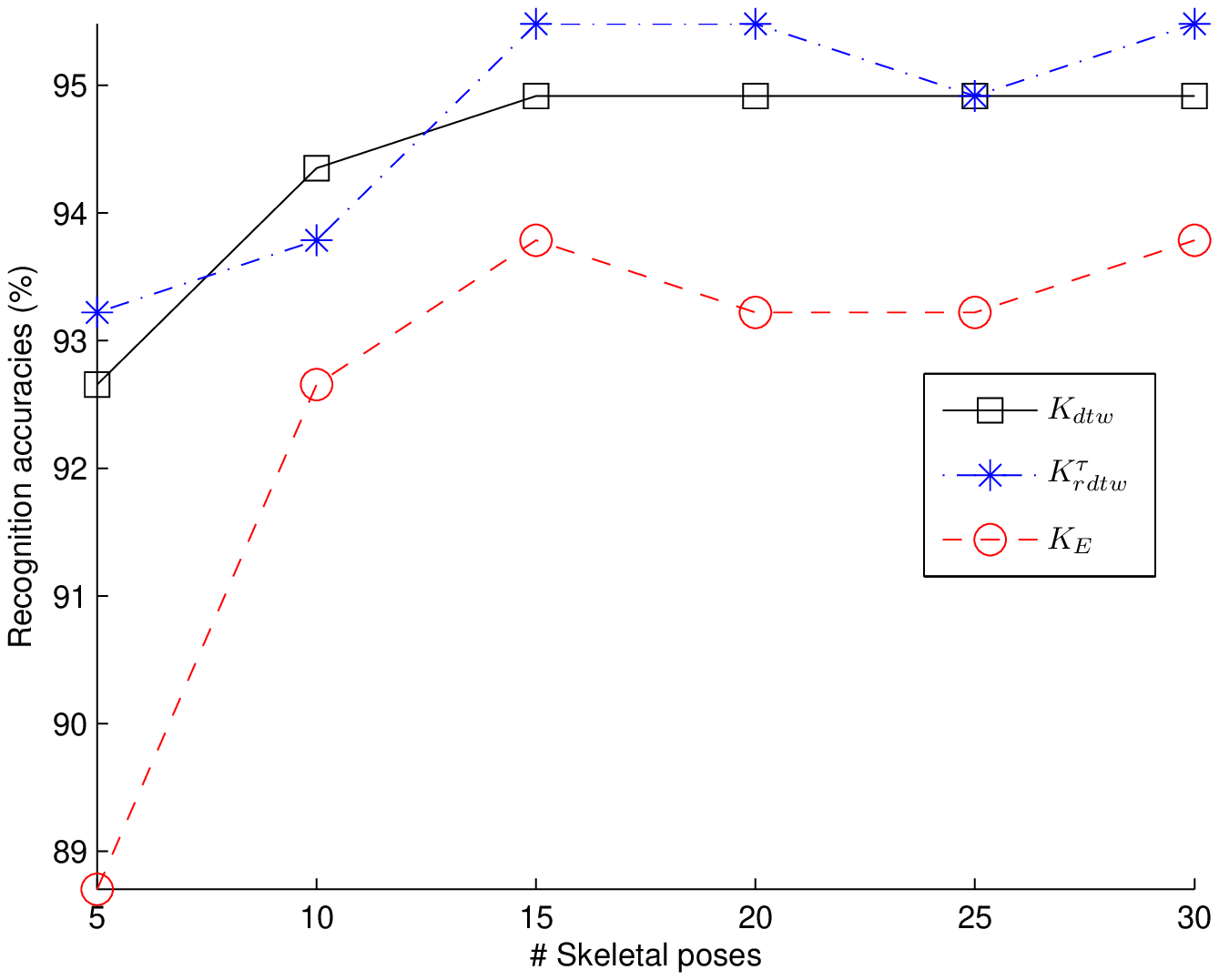} \\
    \includegraphics[scale=.55]{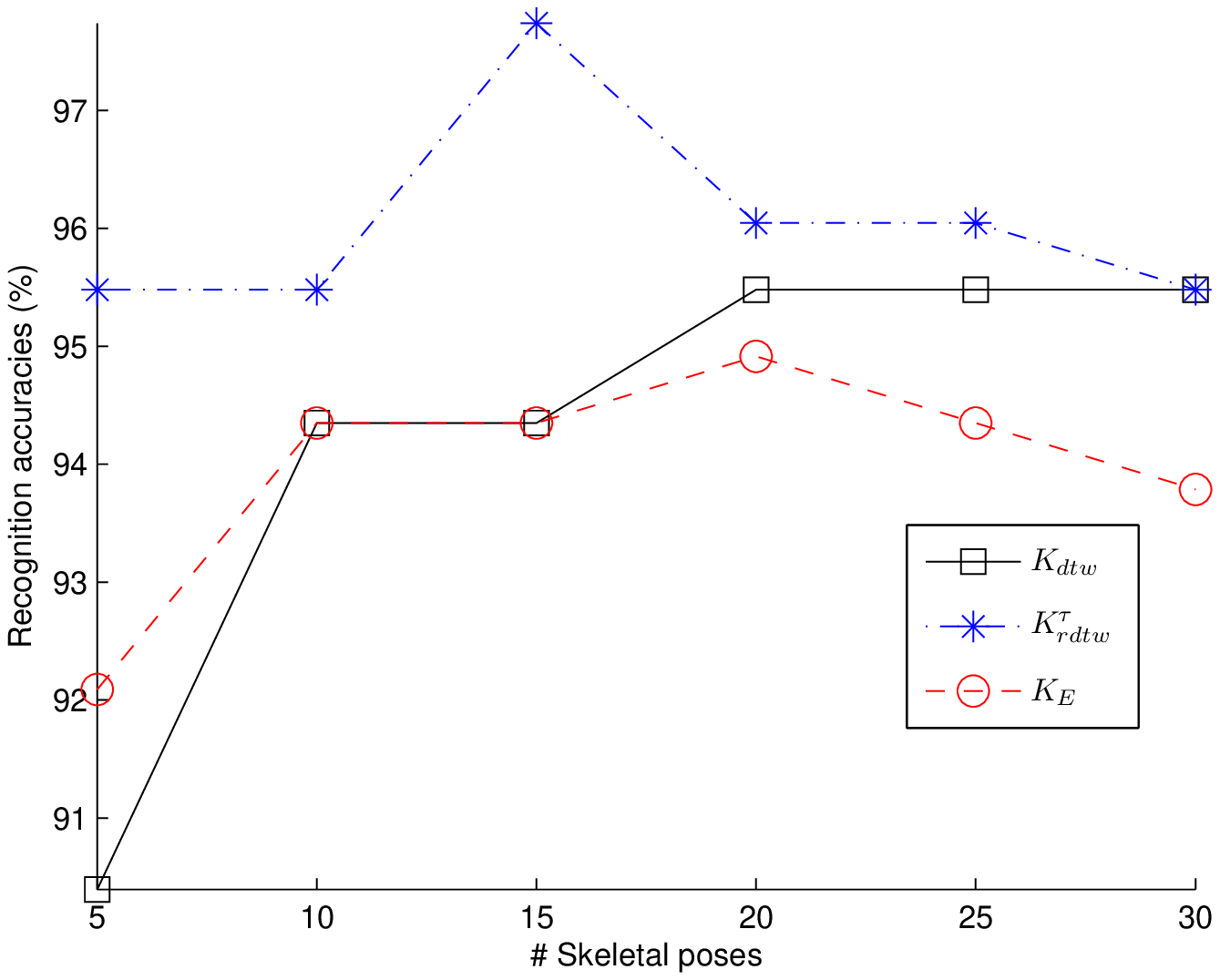}
  \end{tabular}
  \caption{Classification accuracies for HDM05-2 task 
  as defined in \cite{OfliF2013} 
  uniform down-sampling, full body (top), adaptive down-sampling, full body (middle), adaptive down-sampling, end effector extremities (bottom), when the number of skeletal poses varies: $K_{E}$ (red, circle, dash),  $K_{dtw}$ (black, square, plain), $K_{rdtw}^\tau$ (blue, star, dotted).}
\label{fig:HDM05Ofli2013}
\end{figure}

\begin{figure}[H]
  \centering
  \begin{tabular}{cc}
	\includegraphics[scale=.55]{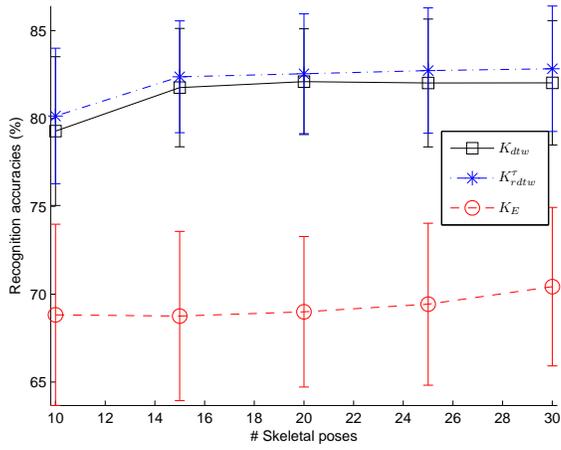} \\
    \includegraphics[scale=.55]{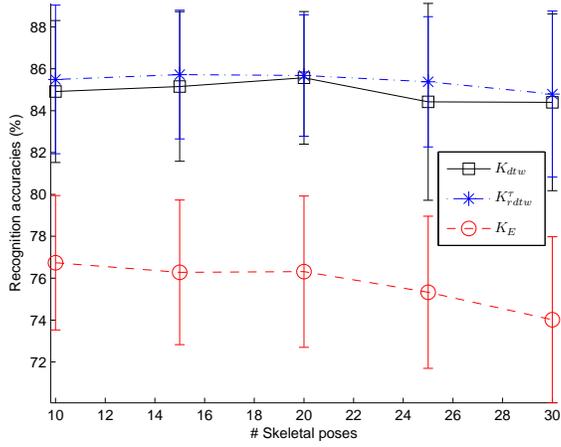} \\
    \includegraphics[scale=.55]{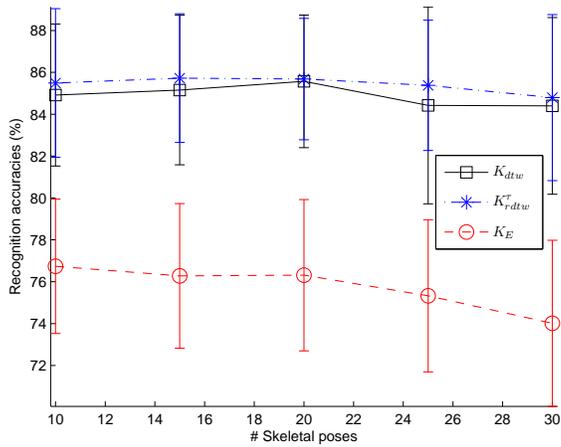}
  \end{tabular}
  \caption{Classification accuracies for the  MSRAction3D data set, uniform down-sampling, full body (top), adaptive down-sampling, full body (middle), adaptive down-sampling, end effector extremities (bottom), when the number of skeletal poses varies: $K_{E}$ (red, circle, dash),  $K_{dtw}$ (black, square, plain), $K_{rdtw}^\tau$ (blue, star, dotted). Additionally, the cross validation on subjects (252 tests) allows to show the variance of the results. }
\label{fig:MSR}
\end{figure}

\subsection{Results and analysis}

For the three considered tasks, we present the results obtained using a SVM classifier built from the LIBSVM library \cite{Libsvm01}, the elastic non definite kernel $K_ {dtw}= e^{-d_{dtw}(.,.)/(2.\sigma^2)}$, the elastic definite kernel $K_{rdtw}^\tau$, and as a baseline, the Euclidean RBF kernel, $K_ {E} = e^{-d_{E}^2(.,.)/(2.\sigma^2)}$. \\

Figures \ref{fig:HDM05Ofli2012}, \ref{fig:HDM05Ofli2013} and \ref{fig:MSR} present the classification accuracies for respectively the HDM05-1, HDM05-2 and MSRAction3D tasks for the test data when the number of skeletal postures selected after down-sampling varies between 5 and 30. For these three figures, the top sub-figure presents classification accuracies when the FBD (Full Body) descriptor associated to a uniform down sampling (UDS) are used, the middle sub-figure  classification accuracies when the FBD (Full Body) descriptor associated to an adaptive down sampling (ADS) are used, and the bottom sub-figure gives classification accuracies when the EED (End Extremities) descriptor associated to an adaptive down sampling (ADS) is used.


On all three figures, we observe that the down-sampling does not degrade the classification accuracies when the number of poses is over 10, and for some tasks it may even significantly improves the accuracy. This is likely due to the noise filtering effect of the down-sampling. High levels of down-sampling (e.g. when 10 to 15 postures are retained to describe movements, which represents an average compression ratio of 97 \% for  HDM05 and 70 \% on MSRAction3D) lead to very satisfactory results (90 to 98 \% for the two HDM05 tasks: Figures \ref{fig:HDM05Ofli2012} and \ref{fig:HDM05Ofli2013},  and 82 to 86 \% for the MSRAction3D task: \ref{fig:MSR}). Best results are obtained for a number of skeletal poses between 15 and 20, when the EED descriptor is used in conjunction with an adaptive down sampling. The SVM classifier constructed on the basis of the regularized kernel $K_{rdtw}^\tau$  produces the best recognition rates (>= 96 \% for the two HDM05 tasks). We note that the MSRAction3D task is much more difficult since it consists in a cross validation based on the performing subjects.  Much lower performance are obtained for the SVM built on the basis of the Euclidean distance; in addition, if very good classification rate (96 \%) is obtained on the training data, due to the noisy nature of Kinect data and the inter subject variability, the recognition rate on the test data falls down from 82 to 86 \% .

\begin{table*}[!ht]
 \begin{center}
   \tabcolsep = 2\tabcolsep
   \begin{tabular}{lcccccc}
   \hline\hline
     FBD-UDS           & $K_{E}$ (L) & $K_{E}$ (T) &   $K_{dtw}$ (L) & $K_{dtw}$ (T) & $K_{rdtw}^\tau$ (L) & $K_{rdtw}^\tau$ (T)\\
   \hline
   Mean & 87,71	& 69,73 & 96,04	& 81,41	& 96,65	& 82,50        \\
   Stand. dev. & 2,34	& 5,73 & 1,36 & 5,04 & 1,13	 &3,22	 \\
   \hline
   \hline
   FBD-ADS & $K_{E}$ (L) & $K_{E}$ (T) &   $K_{dtw}$ (L) & $K_{dtw}$ (T) & $K_{rdtw}^\tau$ (L) & $K_{rdtw}^\tau$ (T)\\
   \hline
   Mean        & 85,06	& 76,48 & 92,13	& 84,72	& 91,89	& 85,09        \\
   Stand. dev. & 3,01	& 3,18 & 2,74 & 3,31 & 2,66	 &3,58	 \\
   \hline
   \hline
   EED-ADS & $K_{E}$ (L) & $K_{E}$ (T) &   $K_{dtw}$ (L) & $K_{dtw}$ (T) & $K_{rdtw}^\tau$ (L) & $K_{rdtw}^\tau$ (T)\\
   \hline
   Mean & 92.46	& 76.27 & 97.14	& 85.16	& \textbf{97.19}	& {85.72}        \\
   Stand. dev. & 3.45	& 0.96 & 	3.57 & 5,04 & 0.93	 &3.07	 \\
   \hline
   \end{tabular}
\caption{Means and standard deviations of classification accuracies on the MSRAction3D data set obtained according to a cross-validation on the subjects (5-5 splits, 252 tests) (L): on the training data, (T): on test data for a number of skeletal postures equal to 15.} \label{tab:resMSRAction3D}
 \end{center}
\end{table*}

Table \ref{tab:resMSRAction3D} gives for the MSRAction3D data set and for the SVM based on $K_{E}, K_{dtw}$ and $K_{rdtw}^\tau$ kernels, means and standard deviations, obtained on the training data (L) and testing data (T), of recognition rates (classification accuracies)  when performing the cross-validation over the 10 subjects (5 train - 5 test splits leading to 252 tests), for the full body descriptor (FBD) and the end extremities descriptor (EED) associated either to a uniform down-sampling (UDS) or an adaptive Down Sampling (ADS). For this test, movements are represented as sequences of 15 skeletal postures. The drop of accuracies between learning and testing phases is due, on this dataset,  to the large inter subjects variability of movement performances. 
Nevertheless, our experiment shows that the best average classification accuracies (obtained in general with minimal variance) are obtained for the most compact movement representation, i.e. when the EED descriptor is used associated to an adaptive down-sampling. This is true both for the training and testing datasets. \\

For comparison, table \ref{tab:resComp} gives results obtained by different methods of the state-of-the-art and compare them with the performance of an SVM that exploits the regularized DTW kernel ($K_{rdtw}^\tau$) associated to the end extremity descriptor (EED) and an adaptive down-sampling (ADS) of 15 skeletal poses. To that end, we have reimplemented the Cov3DJ approach \cite{Hussein2013} to get, for the MSRAction3D data set, the average result given by a 5-5 cross-validation on the subjects (252 tests). This comparative analysis shows that the SVM constructed from regularized DTW kernel provides results slightly above the current state-of-the-art for the considered data sets and tasks.

\begin{table}[ht]
 \begin{center}
   \tabcolsep = 2\tabcolsep
   \begin{tabular}{lc}
   \hline\hline
  HDM05-1 &  Accuracy (\%)\\
   \hline
SMIJ \cite{OfliF2012} & 84.40\\
Cov3DJ, L = 3 \cite{Hussein2013} & 95.41\\
$SVM-K_{rdtw}^\tau$, EED, ADS with 15 poses & \textbf{96.33} \\
   \hline
   \hline
     HDM05-2 &  Accuracy (\%)\\
   \hline
SMIJ \cite{OfliF2013}, 1-NN & 91.53 \\
SMIJ \cite{OfliF2013}, SVM & 89.27 \\
$SVM K_{rdtw}^\tau$, EED, ADS with 15 poses & \textbf{97.74}\\
\hline\hline
MSR-Action3D &  Accuracy (\%)\\
\hline
Cov3DJ, L=3 \cite{Hussein2013} & $72.33 \pm 3.69$ \footnotemark[2]\\
HON4D, \cite{Oreifej2013}, & $82.15 \pm 4.18$\\
$SVM K_{rdtw}^\tau$ 15 poses, & \textbf{85.72 $\pm$ 3.07}\\
\hline
   \end{tabular}
\caption{Comparative study on the MSRAction3D dataset,  according to a cross-validation on the subjects (5-5 splits, 252 tests).} 
\label{tab:resComp}
 \end{center}
\end{table}
\footnotetext[2]{according to our own implementation of Cov3DJ}

\begin{figure}[ht]
  \centering
  \begin{tabular}{cc}
	\includegraphics[width=75mm]{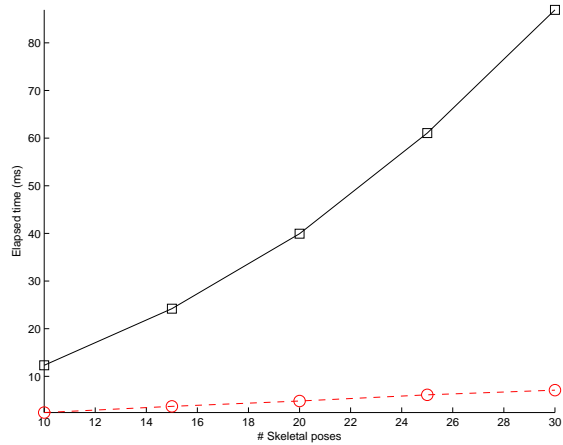}\\
  \end{tabular}
  \caption{Elapsed time as a function of the number of skeletal poses (10 to 30 poses): i) RBF Euclidean Kernel, Red/round/dashed line, ii) RBF DTW kernel, Black/square/plain line, iii)  normalized regularized DTW kernel($K_{rdtw}^\tau$), blue/star/dotted line. }
\label{fig:elapsedTime}
\end{figure}
Finally, in Figure \ref{fig:elapsedTime}, we give the average CPU elapsed time for the processing of a single gestural MSRAction3D action when varying the number of retained skeletal poses. The test has been performed on an Intel Core i7-4800MQ CPU, 2.70GHz. Although the computational cost for the elastic kernel is quadratic, the latency for the classification of a single gestural action using a SVM-$K_{rdtw}^\tau$ is less than 25 milliseconds when 15 poses are considered, which effectively meets easily \textit{real-time} requirements. 

\section{Conclusion and perspectives}

In the context of isolated action recognition, where few studies explicitly consider dimension reduction along both the spatial and time axes simultaneously, we have presented a recognition model based on the dimensionality reduction of the skeletal pose descriptor and the down-sampling of motion sequences coupled to elastic kernel machines. Two ways of down-sampling have been considered: a uniform down-sampling that evenly selects samples along the time axis and an adaptive down-sampling based on a piecewise linear approximation model. The dimensionality reduction of the skeletal pose descriptor is straightforwardly obtained by considering only end effector trajectories, which is consistent with some sensorimotor perceptual evidence about the way human beings perceive and interpret motion. On the data sets and tasks that we have addressed, we have shown that, even when quite important down-sampling is considered,  the recognition accuracy only slightly degrades. In any case, best accuracies are obtained when an adaptive down-sampling is used on the end effector 3D trajectories. The temporal redundancy is therefore high and apparently not critical for the discrimination of the selected movements and tasks. In return, the down-sampling benefits in terms of computational complexity is quadratic with the reduction of the number of skeletal postures kept along the time axis.  

Furthermore, the elasticity of the kernel provides a significant performance gain (comparatively to kernel based on the Euclidean distance) which is very important when the data are characterized by high variability. Our results show that a SVM based on a regularized DTW kernel is very competitive comparatively to the state-of-the-art methods applied on the two tested data sets, even when the dimension reduction on the time axis is important. The down-sampling and dimensionality reduction of the descriptor ensures that this approach meets the real-time constraint of gesture recognition. 

This study opens perspectives to the use of elastic kernels constructed from more sophisticated time elastic distances~\cite{MarteauTWED09} that cope explicitly with time stamped data, associated to adaptive sampling techniques such the one developed in this paper or more sophisticated techniques capable of extracting the most significant and discriminant skeletal poses in movement sequences, based on semantic segmentation.  
We also aim at testing these powerful tools to more complex tasks, where skilled gestures are studied, or/and expressive variations are considered.



\bibliographystyle{apalike}
\bibliography{biblio}


\end{document}